\documentclass[a4paper]{article}

\usepackage{INTERSPEECH2019}
\usepackage[hyphens,spaces,obeyspaces]{url}
\usepackage{comment}
\usepackage{subcaption}
\usepackage{xcolor}

\renewcommand{\vec}{\mathbf} 

\usepackage[normalem]{ulem}

\title{Towards Debugging Deep Neural Networks by Generating Speech Utterances}
\name{Bilal Soomro, Anssi Kanervisto, Trung Ngo Trong, Ville Hautam\"aki}
\address{
  School of Computing, University of Eastern Finland, Joensuu, Finland
}\email{\{bilals, anssk, trung, villeh\}@uef.fi}

\begin{document}
\maketitle
\begin{abstract}

    Deep neural networks (DNN) are able to successfully process and classify speech utterances. However, understanding the reason behind a classification by DNN is difficult. One such debugging method used with image classification DNNs is \textit{activation maximization}, which generates example-images that are classified as one of the classes. In this work, we evaluate applicability of this method to speech utterance classifiers as the means to understanding what DNN "listens to". We trained a classifier using the speech command corpus and then use activation maximization to pull samples from the trained model. Then we synthesize audio from features using WaveNet vocoder for subjective analysis. We measure the quality of generated samples by objective measurements and crowd-sourced human evaluations. Results show that when combined with the prior of natural speech, activation maximization can be used to generate examples of different classes. Based on these results, activation maximization can be used to start opening up the DNN black-box in speech tasks.  
\end{abstract}

\section{Introduction}
    DNNs have 
    produced dramatic improvements over the previous baseline, by the combination of the increase of computing power, huge datasets and algorithmic tweaks~\cite{Goodfellow-et-al-2016}.  Deep models are widely used in speech applications and have shown state of the art results in various speech tasks \cite{deep-neural-networks-for-acoustic-modeling-in-speech-recognition, sainath2015convolutional, sainath2015deep}. This success has led researchers to investigate how the inner workings of neural networks behave so they can be analyzed and further improved. Although DNNs have shown to perform exceptionally well in classification tasks, it has proven to be difficult to peek inside the black box  \cite{Inceptionism45507}.
    
    Classical recognition models were based on the understanding that the complete processing pipeline can be split into disjoint tasks that are then separately optimized,  
    i.e. splitting to a separate feature extraction and classification sub-tasks. Conventional ideas in how to split then depends a lot on the mental models that we have, sometimes to the detriment of the performance~\cite{Sutton2019}.  
    End-to-end (E2E) training and models have come to change all that~\cite{Graves2006}. Those models allow training of all parameters of the pipeline using the final loss function. It allows researchers to find out where our mental models were faulty, such as in E2E dialogue act recognition it was found out that underlying ASR component did not need to be highly optimized~\cite{Serdyuk2018}. In the case of language recognition, E2E models have been successful if acoustical conditions in training and evaluation sets are close~\cite{Trong2018}. However, it has been found in multiple studies that intermediate embedding approach  (i-vector~\cite{Dehak10frontend} or x-vector~\cite{snyder2018vector}) is more robust in case conditions differ between train and eval sets~\cite{Trong2016}.  Thus, it is clear that the practitioners developing these models need better debugging techniques.  
    
    Some debugging techniques have been already commonly used in the vision community, such as \textit{deep visualizations}~\cite{nguyen2016multifaceted}. It involves taking a trained DNN and creating synthetic images that produce specific neural activations of interest.  In addition, researchers have tried to examine various methods to extract what the models have learned in deep CNNs \cite{nguyen2016multifaceted, dosovitskiy2016inverting, mahendran2016visualizing}. Similar CNN debugging strategies have been utilized in the raw waveform modeling~\cite{Sainath2017}. In that work, it was found that the learned filters center frequencies in the stereo microphone system were the same but the spatial responses were different.  
    
    
    Nguyen, \textit{et al.} \cite{nguyen2016synthesizing} used activation maximization to visualize what a CNN has learned. They demonstrated that visualizing the input layers of the neural network did not produce realistic images and does not help researchers understand clearly what the network has learned. The use of adversarial examples~\cite{Goodfellow2015} is an analogous idea, where instead of finding an input that maximizes the output activation to a correct class, we find how to modify the input so that classifier is fooled. What is generally observed is that adversarial noise, additive in the signal domain, appears visually to be random. In general, it does not inform of the properties of the classifier. But, what if we have a \textit{deep generator network} (DGN) trained on the signal domain, such as images in or speech features. Then the question of debugging can be formed in such a way that we can change the latent representation to generate a signal that maximizes the output activation. This allows us to produce more realistic images that help us to understand and debug what the network has learned in a more interpretative way \cite{nguyen2016synthesizing}.
    
    
  \begin{figure}[ht!]
  \centering
  \includegraphics[width=0.88\linewidth]{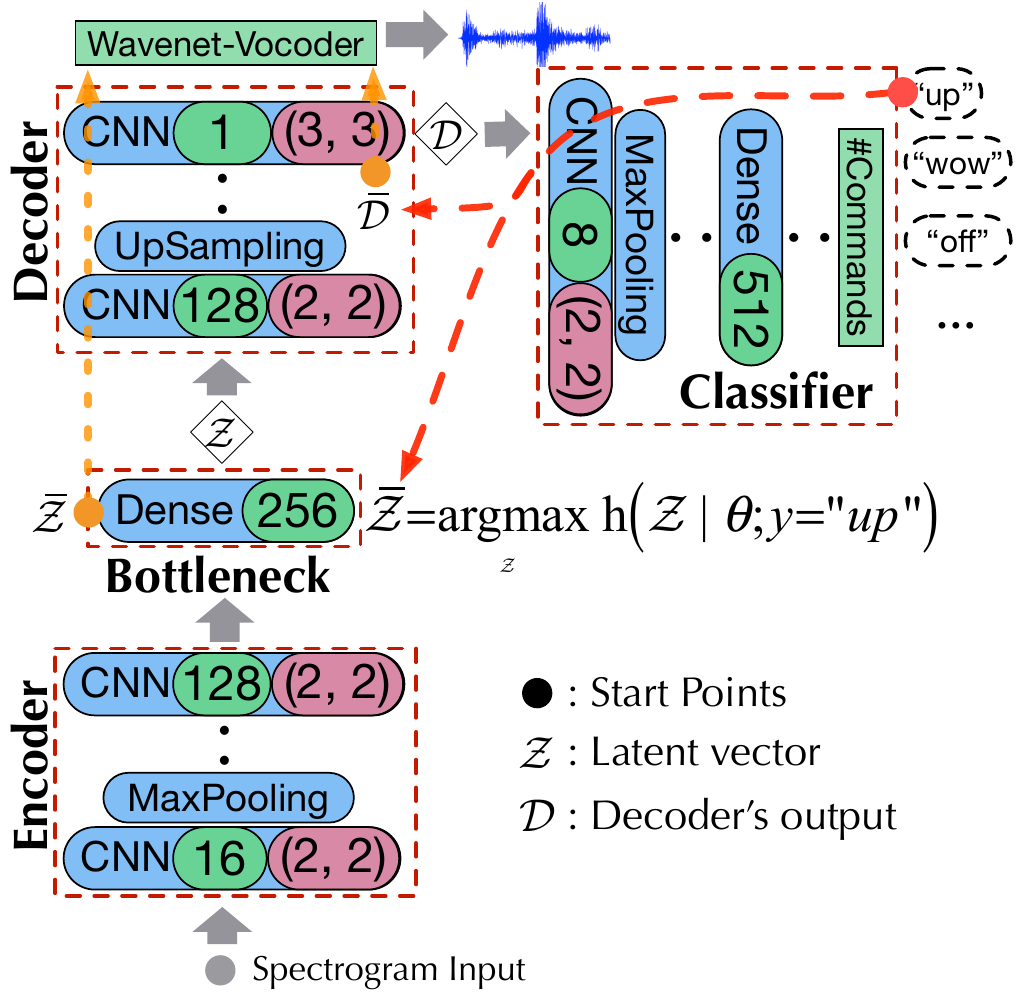}
  \caption{The design of our neural networks. The flow of the information during training phase is represented by the grey arrow. The backpropagated information during the input maximization phase is illustrated by the dashed red line. The debugging phase, when we manipulate the code vector for specific pattern, is represented by the dotted orange line.}
  \label{fig:network_designs}
  \end{figure}
  
    In this work, we apply this technique to understand the inner workings of the speech commands classifier. We use DGN with activation maximization to produce speech utterances which classify strongly to a target class. We then evaluate the results using objective and perceptual tests (MTurk). We use the technique, in a proof of concept fashion, to debug misclassified test set utterances. 
    
    

\section{Sampling from a trained classifier}
    \subsection{Activation maximization}

        Activation maximization is the task of finding input patterns which maximize the activation of a given unit \cite{erhan2009visualizing}. This itself is an optimization problem. Let $\theta$ be fixed neural network parameters and $h_i(\vec{x} ; \theta)$ the activation of the neuron $i$ and $\vec{x}$ is the input of the neural network. The whole neural architecture is then implicitly included in the function $h_i$ using fixed parameters $\theta$. Input $\vec{x}^*$ that maximizes the activation $h_i$ is then 
        \begin{equation}
            \vec{x}^* := \arg\max_{\vec{x}} h_i(\vec{x} ; \theta).
        \end{equation}
        Since the neural network is differentiable, we can apply gradient descent with learning rate $\alpha$ to obtain a local maximum of $h_i$ around some starting position $\vec{x}$ by repeating
        \begin{equation}
            \vec{x}_{n+1} := \vec{x}_n + \alpha \nabla_{\vec{x}_n} h_i(\vec{x}_n ; \theta) 
        \end{equation}
        until desired results or after enough iterations. After the  process has ended, we can take $\vec{x} - \hat{\vec{x}} = \vec{x}_{\mathrm{diff}}$, where $\hat{\vec{x}}$  is the result of the activation maximization. This process can be viewed as generating an additive ``noise". 
        
        
    
    \subsection{Generating classifier dependent noise}
        Using the previous method of maximizing an activation of a neuron, we can maximize the output of a classification network to create an example that is ``maximally" considered to be of one class. With image classification, for example, we could try maximizing random noise into class ``chair" and expect to have an image roughly representing a chair. These images could then be used to understand what neural network is ``looking at" when it interprets an image as a chair.
        
        In practice, this alone does not produce desired results~\cite{nguyen2016synthesizing}. Instead of generating images of chairs, the maximization procedure generates seemingly random spots on the image. While the images used in classification come from the distribution of natural images, the neural network still can classify any set of pixels into any class. The activation maximization technique then moves us away from the set of natural images into unnatural area, where the image can contain random pixels. Instead, we wish to remain in the set of natural images.
        
    \subsection{Using deep generative network}
        Nguyen \textit{et al.} \cite{nguyen2016synthesizing} propose using a decoder trained to generate natural images from a lower dimensional latent code. Such decoder can be obtained by e.g. training an auto-encoder on a dataset of natural images, and keeping the decoder part. Nguyen \textit{et al.} then use this decoder to turn a latent code into an image, and then apply activation maximization on the latent. Since this decoder is only trained to create natural imagery, it will limit the image to be in the set of natural images. With this method, the authors were able to create natural imagery of what neural networks learned to see. 
        
        In this work, we evaluate the use of activation maximization alone and activation maximization with such a prior in speech classification. We do this by evaluating generated samples with objective measures and human evaluations.
        
\section{Experimental setup}
\label{sec: setup}

\subsection{Dataset}
We use the Speech Commands corpus v0.02 \cite{speechcommandsv2} for the experiments. It contains $105,829$ utterances recorded from $2,618$ speakers. The utterances contain $35$ commands in which there are twenty trigger words. The corpus also contains words that sound similar to the core words such as "Tree" and "Three", which adds some challenge for the classification models. The recording environment of the speech utterances in the corpus vary in quality to mimic real world environments and different devices. The v0.01 of the Speech Commands corpus was featured in the TensorFlow speech recognition challenge on Kaggle \cite{speech_commands_v1}. The winner of that competition was able to achieve a classification accuracy score of $91\%$.

\subsection{WaveNet Vocoder}

As we wanted to listen to what the speech classifier had learned, we needed to synthesize the speech features back into audio. Currently, the state of the art speech synthesizer is the one designed by the DeepMind team called \textit{WaveNet} \cite{wavenet_deepmind}. WaveNet is an audio generative model based on the PixelCNN architecture \cite{van2016wavenet}. It is able to produce the most natural sounding human voice samples and has been deployed at production level, such as in Google's Voice assistant \cite{van2016wavenet}.

To synthesize audio from speech features, we used a pre-trained WaveNet model\footnote{\url{https://github.com/r9y9/wavenet_vocoder}} which was trained on the LJSpeech corpus \cite{ljspeech17}. The WaveNet model was trained using mel-spectrogram features. We used the same feature extraction code provided in the github repository on our speech commands corpus and trained our speech classifier. As a result, we did not have to train our own WaveNet model.

\subsection{Training Speech classifier}
We extracted mel-spectrogram features from the Speech Commands corpus and used a standard CNN model for our speech commands classifier. Although CNNs are popular for machine vision tasks, they have proven to be successful in speech recognition tasks \cite{sainath2015convolutional, sainath2015deep}. Figure \ref{fig:network_designs} shows the structure of the CNN used in our speech classifier model. The classifier was trained for $50$ epochs, and the model achieved an accuracy score of $82.75\%$ in our test set.

\subsection{Prior of speech with auto-encoders}
\label{subsec:prior_with_autoencoder}
For our maximization experiments with a prior, we trained an auto-encoder using the speech commands corpus. The decoder part is then combined with our speech classifier model. This enabled us to maximize the latent codes to generate speech features that activate our classifier's target classes. Figure \ref{fig:network_designs} shows the architecture of the decoder used in our activation maximization experiments. The bottleneck layer in our auto-encoder has no activation function.

\subsection{Activation maximization}
To understand what the DNN ``listens to" in speech recognition tasks, we setup two sets of activation maximization experiments to generate samples for synthesis: \textit{noise-to-class} and \textit{class-to-class}. Both sets of experiments were performed on the classifier model and the combined model.

In the noise-to-class experiment with only the classifier, we generated noise features and then used activation maximization to modify the features to the desired target classes. For our combined model, we generated random latent codes which were then maximized until they produced features which activated the target classes. 

In the class-to-class experiment with only the classifier, we randomly picked speech features from our test set and maximized them to their respective classes. For our combined model, we encoded our test set speech features to latent codes and maximized them to their respective classes, essentially enhancing the classification.

We also used a separate speech classifier model, trained with the exact same specifications as our original speech classifier, to evaluate the resulting speech features. The purpose of this was to determine if the maximized speech features can fool a separate classifier, other than the one used to maximize the classes. The code is available on \url{https://github.com/bilalsoomro/debugging-deep-neural-networks}. 

\subsection{Perceptual experiments}
We used Amazon's Mechanical Turk service to get human listeners to rate our synthesized speech features on their quality. The participants were shown the class label of the recording and asked to rate it on a scale of one to five on how clear and audible the sample is. The rating was explained as (1: Bad, 2: Poor, 3: Fair, 4: Good, 5: Excellent), with "Bad" being described as "completely unclear speech". We also provided examples of recordings that match the ratings to give our subjects a good idea on how to rate them.

There were a total of $650$ speech features synthesized for perceptual evaluation. We selected features that were successfully maximized to the target classes. The recordings contained three maximized noise samples to each of our target classes generated from both the classifier model and the combined model. The recordings also contained three original and maximized test samples taken from both the classifier model and the combined model. When setting up the evaluation tests, we asked for five unique subjects to rate each of the recordings which added up to a total of $3,150$ evaluations.

\section{Results}
\label{sec:results}

\begin{figure}[ht]
  \centering
  \includegraphics[width=\linewidth]{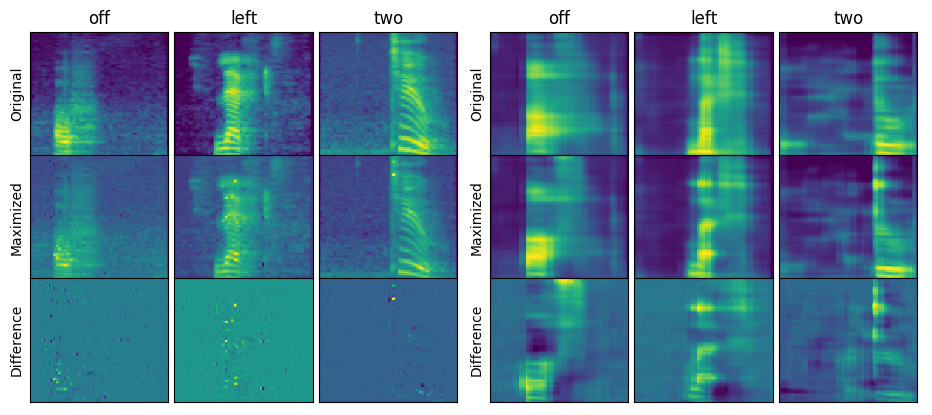}
  \caption{Visualization of speech features of class label "Off", "left" and "two". The left figure shows test speech features maximized using the classifier. The right figure shows test speech features maximized using the combined model of the decoder and speech classifier.}
  \label{fig:test_max}
\end{figure}

\subsection{Objective evaluations with separate classifier}
\begin{figure}[t]
  \centering
  \includegraphics[width=\linewidth]{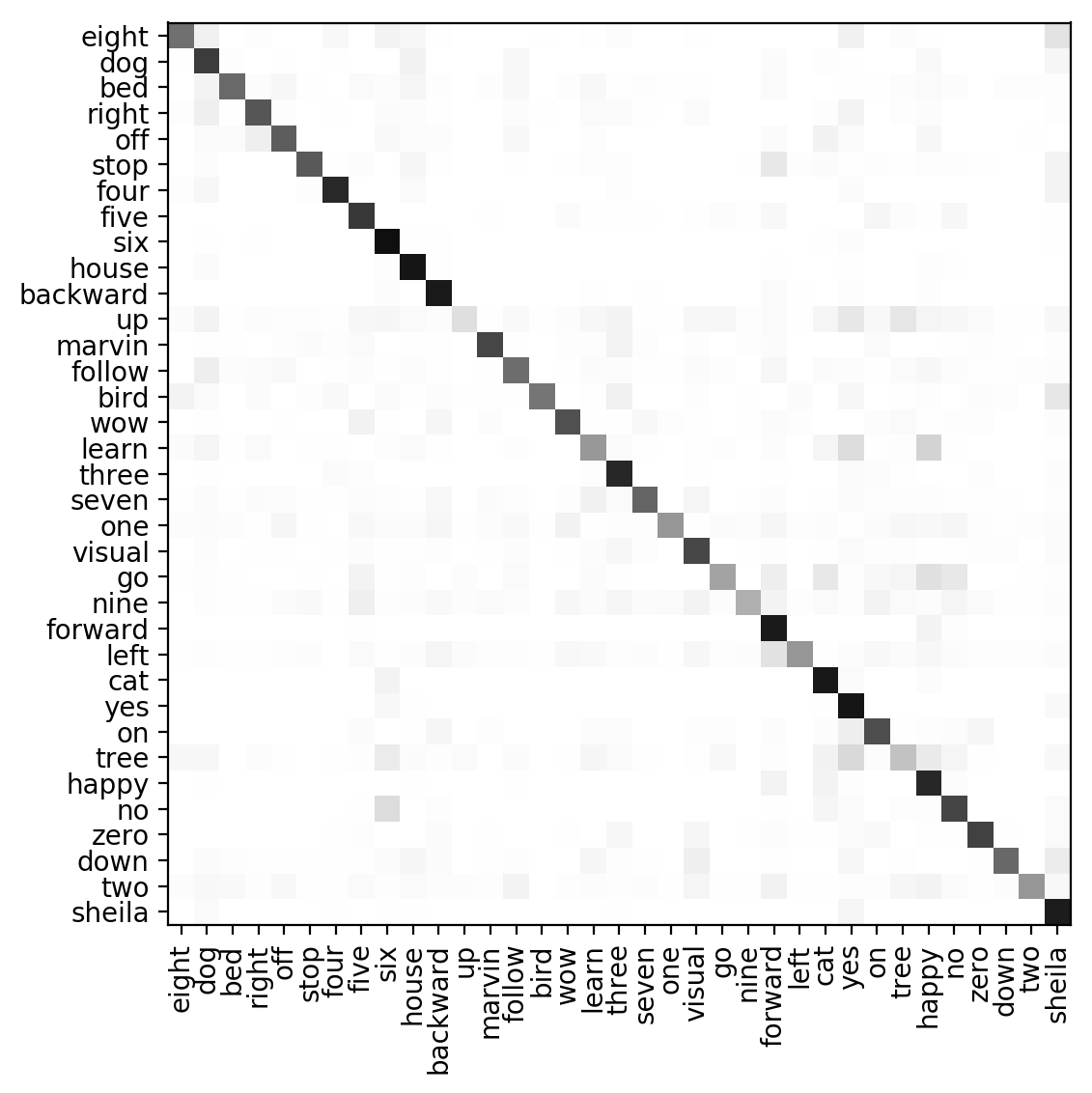}
    \caption{Results of maximizing $10,000$ random latent codes per class (rows) using the decoder setup. Columns represent classification score from a separate classifier, darker being higher score. A perfect generator would have solid black diagonal line. Most classes are maximized correctly with values on diagonal ranging from $0.12$ to $0.92$, with some outlier classes like "up", "tree" and "go" which get maximized to other classes.}
    \label{fig:noise2class}
\end{figure}

We consider maximization successful if original and separate classifier both classify the maximized sample to the target class. To evaluate the performance of the classifier and combined models, we maximized $10,000$ random latent codes / features per class into that class, and classified the maximized results with both classifiers. The maximization procedure worked $96.8\%$ of the time with classifier setup, and $92.1\%$ of time with decoder setup. However, when classifying these samples with the separate classifier, the classifier setup only worked $7.6 \%$ of the time while decoder setup worked $67.5\%$ of the time. The per-class results are shown in Figure \ref{fig:noise2class} for the decoder setup, where we can see the maximization worked for most the classes minus a few outliers.  

By visually inspecting the output features from the maximization (Figure \ref{fig:test_max}), we see that the classifier alone introduces seemingly random pixels to the feature space while still successfully doing the maximization. When combined with a decoder, the maximization introduces patterns reminiscent of formants and general speech structure (horizontal stripes, no individual pixels changed). 

As expected, use of decoder limits modifications by maximization into speech-like structures, while maximization with the classifier alone is free to abuse the full space of possible features. This includes one-pixel changes (when represented as an image), which are unnatural for speech. However, despite decoders restrictions for the modifications, it helps maximization procedure to reach higher activation values.

\subsection{Perceptual evaluations}
\begin{figure}[ht]
  \centering
  \includegraphics[width=\linewidth]{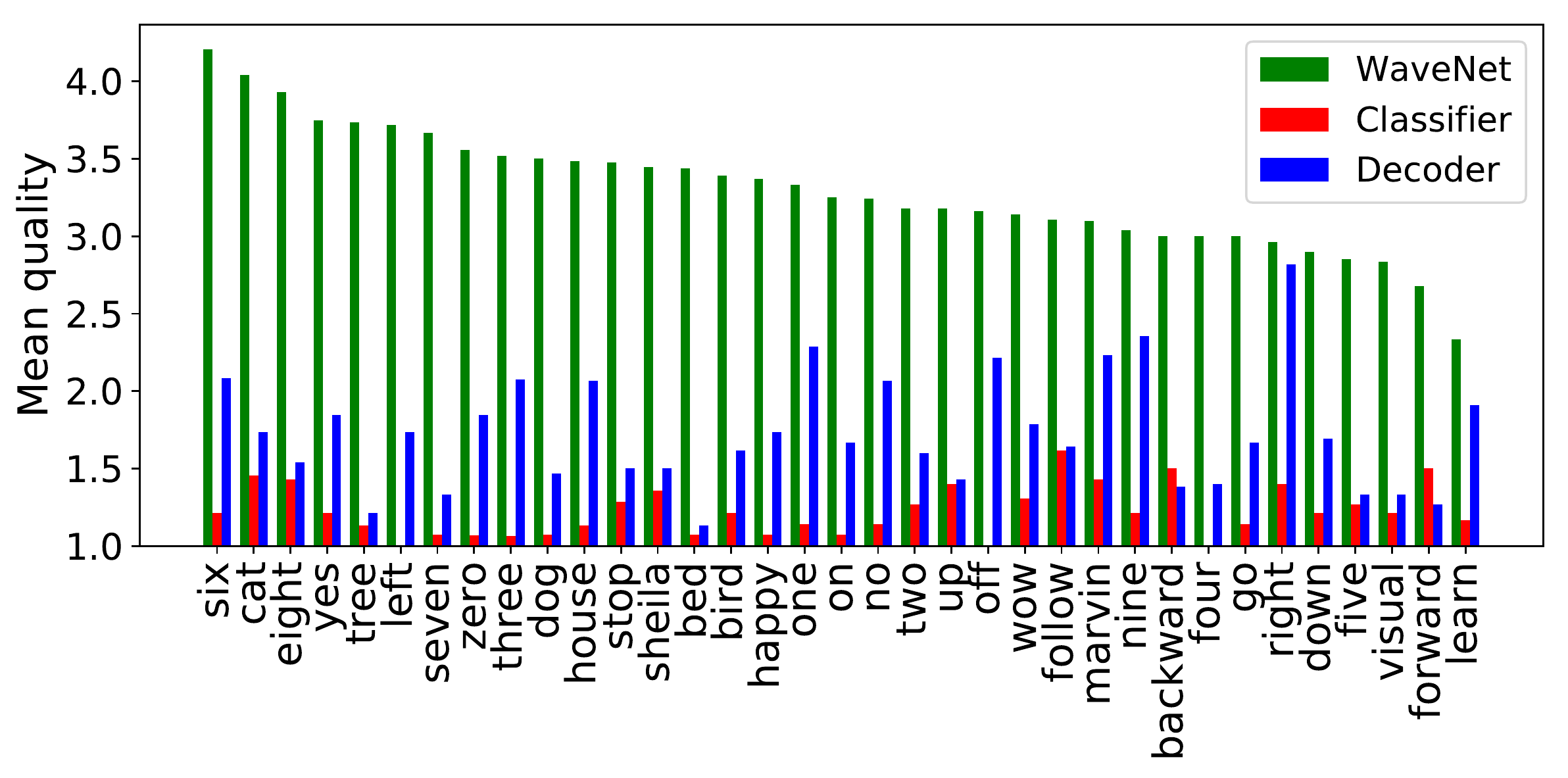}
    \caption{Results of perceptual evaluations on quality of synthesized samples, ranging from one to five and averaged over $\approx 15$ answers. The WaveNet synthesizing alone distorts the samples (green bars well below four). Using decoder produces higher quality samples then classifier alone overall, except for longer commands ("backward", "forward"). This indicates that decoder is able to generate higher quality samples.}
    \label{fig:noise2class_perceptual}
\end{figure}

Figure \ref{fig:noise2class_perceptual} shows results of human evaluations on quality of noise-to-class maximized samples, with and without decoder. Each bar is an average over $15-50$ separate human evaluations. The WaveNet synthesizing alone caps the quality to around $4.0$. Samples generated with decoder maximization constantly have higher average score than classifier alone. With decoder setup, $23/35$ of the classes reach above $1.5$ average quality rating, while with classifier setup only one class reaches this. Note that class one ("Bad") was reserved for samples that did not contain any structures of speech, while class two ("Poor") and above should contain audible speech. 

In class-to-class experiments, the effect was less dramatic but opposite: classifier-maximized samples had on average quality $3.58$ and decoder-maximized samples had $2.97$. This is due to detrimental effect of the encoder-decoder setup which introduces artifacts and especially smoothing in the feature space (see Figure \ref{fig:test_max} for examples of decoded samples). These results indicate that the maximization with the decoder works better, even on the subjective level, with decoder setup being able to generate speech-like samples from random noise more often than classifier setup.

\subsection{Debugging speech processing tasks}

\begin{figure}[t!]
  \centering
  \includegraphics[width=1.03\linewidth]{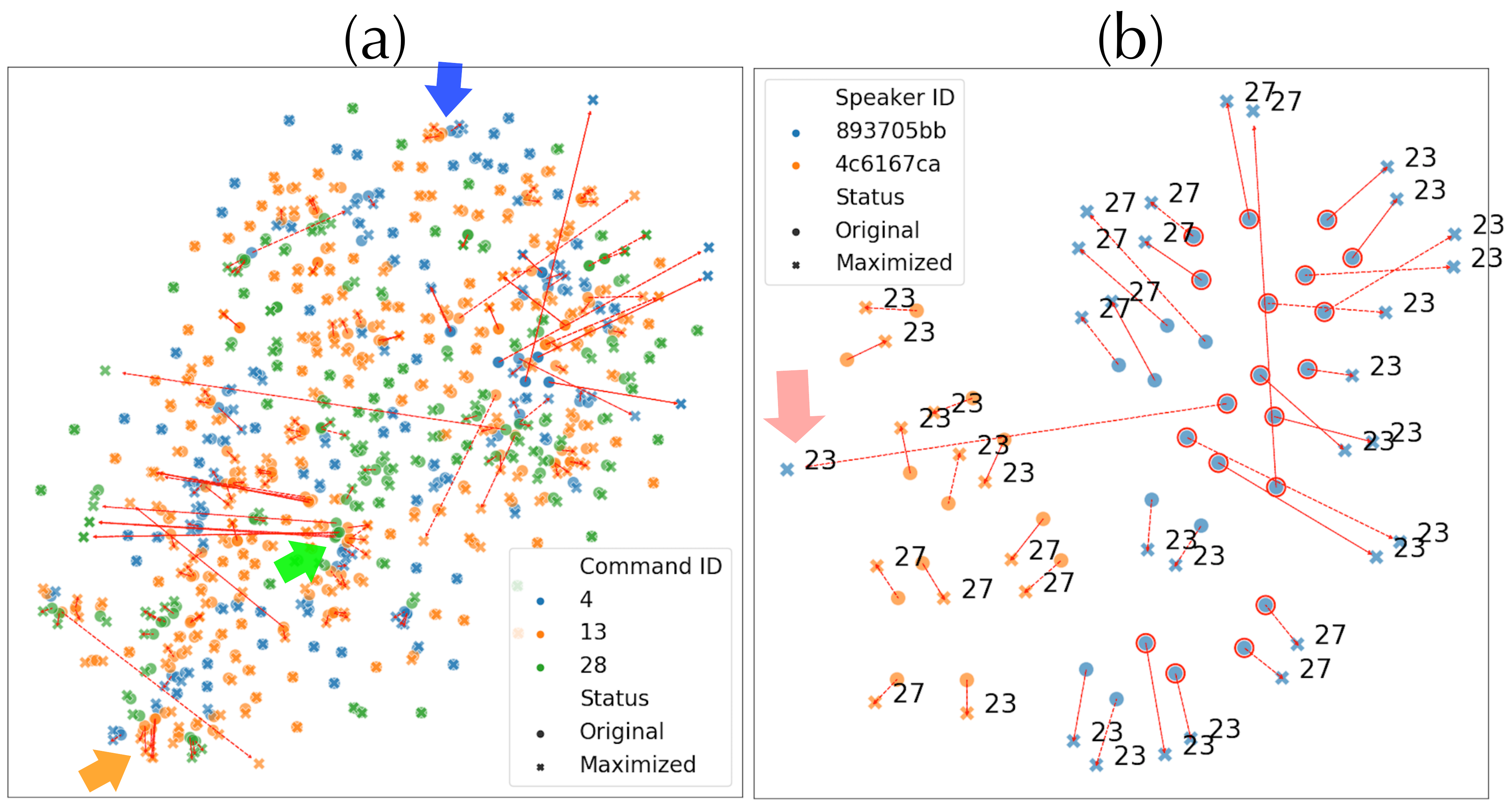}
  \caption{Influence of the maximization process with respect to the command labels (colored) in $\mathrm{(a)}$, and the speaker labels (colored) in $\mathrm{(b)}$. The visualization is obtained by applying t-SNE on the latents. The misclassified cases are highlighted by red circle in $\mathrm{(b)}$.}
  \label{fig:latent_tsne}
\end{figure}

Our speech command classifer is able to obtain $82.75\%$ classification accuracy, we would like to know the reason for misclassification on that $17.25\%$ portion. As our classifier is trained primarily for speech command recognition task, Figure~\ref{fig:latent_tsne}$\mathrm{(a)}$ suggests a strong correlation between the maximized pattern and relevant command information. The algorithm had performed both micro and macro adjustment on the misclassified samples. The green arrow shows that our algorithm moves a cluster of green dots to a separated cluster outside the confusing zone (i.e. the macro adjustment). Additionally, the blue and orange arrow highlight its capability to re-distribute the points within a short distance to remove the confusion between the orange and blue command (i.e. the micro calibration). In general, the maximization process is only activated strongly in the highly confusing area, which is remarkably efficient since many proper clustered points remain unchanged.

We can hypothesize that one of the confounding factors in speech command classifier case is due to speaker variation. As a diagnostic algorithm, our approach could evaluate the effect of speaker variation on the classifier of speech command. Figure~\ref{fig:latent_tsne}$\mathrm{(b)}$ provides strong evidence for the investigation:
\begin{itemize}
    \item The points are clearly pushed away even though coming from the same speaker.
    \item The same command, which annotated by the number, are pushed to the same direction regardless of the speaker ID. An interesting case is highlighted by the red arrow when a data point of the command \textit{23} from the \textit{blue} speaker is \textit{maximized} to group with the $23^{th}$ from the \textit{red} speaker.
\end{itemize}
As a result, we have strong indication that, indeed speaker variation had a strong role in the misclassification, and the issue can be mitigated by further adjusting the classifier (e.g. integrating the pattern extracted from the maximization process).

\section{Conclusions}
\label{sec: conclusions}
In this work, we evaluated maximization activation as a method to ``listen to" what a speech classifier has learned. We performed experiments to maximize random noise to a class and as well as maximizing class to class. Similar to the prior work done on visualizing image classification models, we also observed that performing activation maximization directly on the classifier resulted in unnatural speech features. We observed that the decoder maximization resulted in more natural speech features. We were able to successfully fool a second classifier with features maximized from our combined decoder and classifier model. Our perceptual evaluation results also show that samples from decoder method are subjectively higher. In the future, this work can be applied to more complicated end-to-end classifier tasks i.e. variable length inputs in language identification.

\section{Acknowledgements}
 This research was partially funded by the Academy of Finland (grant \#313970) and  Finnish Scientific Advisory Board for Defence (MATINE) project \#2500M-0106. We gratefully acknowledge the support of NVIDIA Corporation with the donation of the Titan Xp \& V GPUs used for this research.

\bibliographystyle{IEEEtran}
\bibliography{mybib}
\end{document}